\pdfoutput=1
\documentclass[11pt]{article}

\usepackage[final]{acl}

\usepackage{times}
\usepackage{latexsym}
\usepackage[T1]{fontenc}
\usepackage[utf8]{inputenc}
\usepackage{microtype}
\usepackage{inconsolata}
\usepackage{graphicx}
\usepackage{amsmath}
\usepackage{amssymb}
\usepackage{tabularx}
\usepackage{booktabs}
\usepackage{algorithm}
\usepackage{algpseudocode}
\usepackage{algpseudocode}
\usepackage{tikz}
\usepackage{caption} 
\usetikzlibrary{positioning}
\usetikzlibrary{fit}

\newtheorem{prompt}{Prompt}

\title{From Punchlines to Predictions: A Metric to Assess LLM Performance in Identifying Humor in Stand-Up Comedy}

\author{Adrianna Romanowski \\
  International Christian University\\
  \hspace{-0.5cm}\texttt{adaromanowski@gmail.com} \\\And
  Pedro H. V. Valois \\
  University of Tsukuba\\
  \texttt{pedro@cvlab.cs.tsukuba.ac.jp} \\ \And
  Kazuhiro Fukui \\
  University of Tsukuba \\
\hspace{0.7cm}\texttt{kfukui@cs.tsukuba.ac.jp}}

\begin{document}
\maketitle
%%%%%%%%% ABSTRACT
\begin{abstract}
Comedy serves as a profound reflection of the times we live in and is a staple element of human interactions. In light of the widespread adoption of Large Language Models (LLMs), the intersection of humor and AI has become no laughing matter. Advancements in the naturalness of human-computer interaction correlates with improvements in AI systems’ abilities to understand humor. In this study, we assess the ability of models in accurately identifying humorous quotes from a stand-up comedy transcript. Stand-up comedy's unique comedic narratives make it an ideal dataset to improve the overall naturalness of comedic understanding. We propose a novel humor detection metric designed to evaluate LLMs amongst various prompts on their capability to extract humorous punchlines. The metric has a modular structure that offers three different scoring methods -- fuzzy string matching, sentence embedding, and subspace similarity -- to provide an overarching assessment of a model's performance. The model's results are compared against those of human evaluators on the same task. Our metric reveals that regardless of prompt engineering, leading models, ChatGPT, Claude, and DeepSeek, achieve scores of at most 51\% in humor detection. Notably, this performance surpasses that of humans who achieve a score of 41\%. The analysis of human evaluators and LLMs reveals variability in agreement, highlighting the subjectivity inherent in humor and the complexities involved in extracting humorous quotes from live performance transcripts. Code available at \url{https://github.com/swaggirl9000/humor}.
\end{abstract}

\section{Introduction}
Humor plays a significant role in our daily lives and is a fundamental part of human interaction. Despite the rapid advancements in artificial intelligence and human-computer interactions, the field of computational humor lags behind. Improvement in the ability of machines to understand and generate humor has the potential to enhance the naturalness of exchanges with Large Language Models (LLMs). Prior research has demonstrated that humans interact with the personalities of computers similarly to the ways they respond to other humans. As AI systems continue to integrate into e-commerce, virtual reality, and take on personal assistant roles, the necessity for these systems to exhibit a certain level of social intelligence, which goes hand-in-hand with humor, becomes essential~\cite{1613822}.

The tasks of humor detection, evaluation, and generation are consistently a challenge for AI due to humor's reliance on irony, sarcasm, and cultural nuances. Research shows that models trained on diverse datasets, ranging from humorous tweets to funny news headlines to puns, can achieve strong performance on tasks. However, they often struggle with out-of-domain scenarios~\cite{baranov-etal-2023-told} and tend to over rely on stylistic features such as punctuation and question words, rather than a deep semantic understanding~\cite{inacio-etal-2023-humor}. 

\begin{figure}
    \centering
\includegraphics[width=1\linewidth]{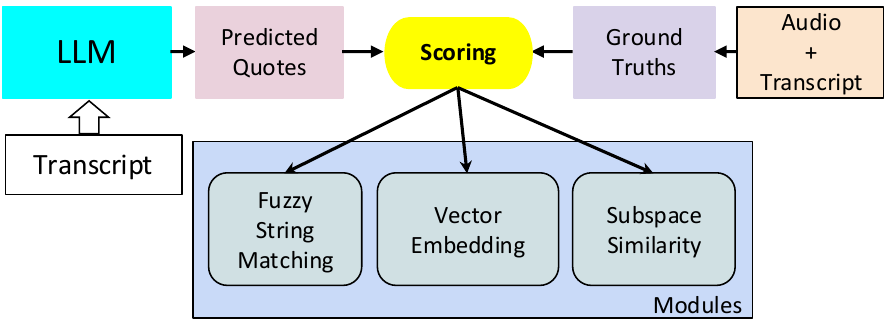}
    \caption{We propose a humor detection metric with three alternative scoring modules -- fuzzy string matching, vector embedding, subspace similarity -- and integrate them to assess a model's predictions with the ground truth, the stand-up comedy quotes that made the audience laugh.}
    \label{fig:short}
\end{figure}

Traditionally, research on humor detection was approached through binary classification tasks, using standalone jokes~\cite{mihalcea-strapparava-2005-making} or occasional jokes within longer presentations~\cite{Hasan2019}.
In this paper, we propose a shift towards using datasets that capture humor within a narrative structure, specifically focusing on stand-up comedy transcripts for humor detection~\cite{mittal2021so, turano-strapparava-2022-making}. Stand-up comedy is a performance where comedians deliver jokes and funny monologues directly to a live audience. Regardless of the diversity in comedic styles, the overarching goal of any comedian remains consistent – to maximize audience laughter – creating a valuable resource for the perception of everyday humor~\cite{Daboin2022-DABWTD-2}. In essence, stand-up comedy serves as both a data source and a pedagogical example for teaching AI the mechanics of humor, especially when the goal is to improve a model’s ability to communicate in a way that feels intuitive and relatable to humans.

LLMs demonstrate notable proficiency across a broad spectrum of tasks, but their performance can fluctuate based on the task's nature. By developing a task-specific metric that focuses on humor detection, we offer a means of evaluation for a nuanced domain like comedy. The simplest method for measuring the capability of a model would be by counting the number of perfect matches. Taking subjectivity into consideration, it is unreasonable to expect perfection, even for humans. Thus, we offer a metric that provides a fair quantitative assessment that encompasses the subjectivity of humor with the probabilistic nature of LLMs. 

Following Figure~\ref{fig:short}, our metric assesses a model’s performance in humor detection in zero-shot prompting scenarios by comparing the similarity of predicted humorous quotes against the ground truth – the punchlines that elicited laughter from the audience. The model operates in a zero-shot setting, meaning it is not provided with examples or prior instructions before prompting. The metric offers a modularized approach with three different ways to output a score. 

First, the most straightforward approach uses fuzzy string matching to compare the similarity of two lists of strings~\cite{inbook},
where each list consists of humorous quotes from a stand-up comedian’s transcript. However, this quick, efficient method does not account for semantics and context, making it too punitive if a model makes a slight mistake when providing the quote. 

Second, the vector embedding module captures semantic similarity between the vector representations of sentences, facilitating a more flexible assessment that emphasizes the underlying meanings of quotes rather than a strict word-for-word correspondence.~\cite{reimers-2019-sentence-bert}. 

Third, while fuzzy string matching and vector similarities offer focused insights on a task, the last module provides a score that reflects the overall capability of a model in humor detection tasks using subspace representations. A subspace is generated for the model after it is prompted with several variations of an instruction and another subspace is generated for the ground truth. The alignment between these two subspaces reflects the structural similarity between the model's outputs and the ground truth for the transcript in a more general way. 

By proposing three distinct scoring modules for assessment, our metric acknowledges the subjectivity of the task, granting the evaluator the flexibility to decide how punitive they want to be towards a model's responses. Fuzzy string matching offers a direct evaluation focusing on precision. Whereas, sentence embeddings are particularly useful when the model generates both a quote and accompanying explanation, allowing for an evaluation of contextual understanding and semantics. Subspaces introduce a novel approach that captures a model's overall ability, considering multiple possible responses for the task in a single score. Balancing these methods gives a well-rounded view of performance, ensuring that both accuracy and deeper semantic understanding are taken into account.

We employ this metric to evaluate the efficacy of several different prompts and various language models. Additionally, we conduct a human evaluation on the same dataset to provide a reliable comparison for model performance. The human-based assessment accounts for the inherent subjectivity of humor, offering a reliable context to gauge the relative performance of the problem at hand. 

The main contributions of this paper are: 
\begin{enumerate}
\setlength{\itemsep}{0pt} 
\setlength{\parskip}{0pt}
\item Introduce a flexible metric that is designed to consider the subjectivity of humor detection tasks, providing a fair measure for the performance of LLMs;
\item Assess the metric across various models and multiple prompt variations, applied to stand-up comedy transcripts;
\item Provide a quantitative assessment of human performance on the same humor detection task, alongside a calculation of agreement ratios between human and LLM-based humor detection, offering a basis for comparison.
\end{enumerate}

\begin{figure*}[!t]
    \centering
    \includegraphics[width=0.8\textwidth]{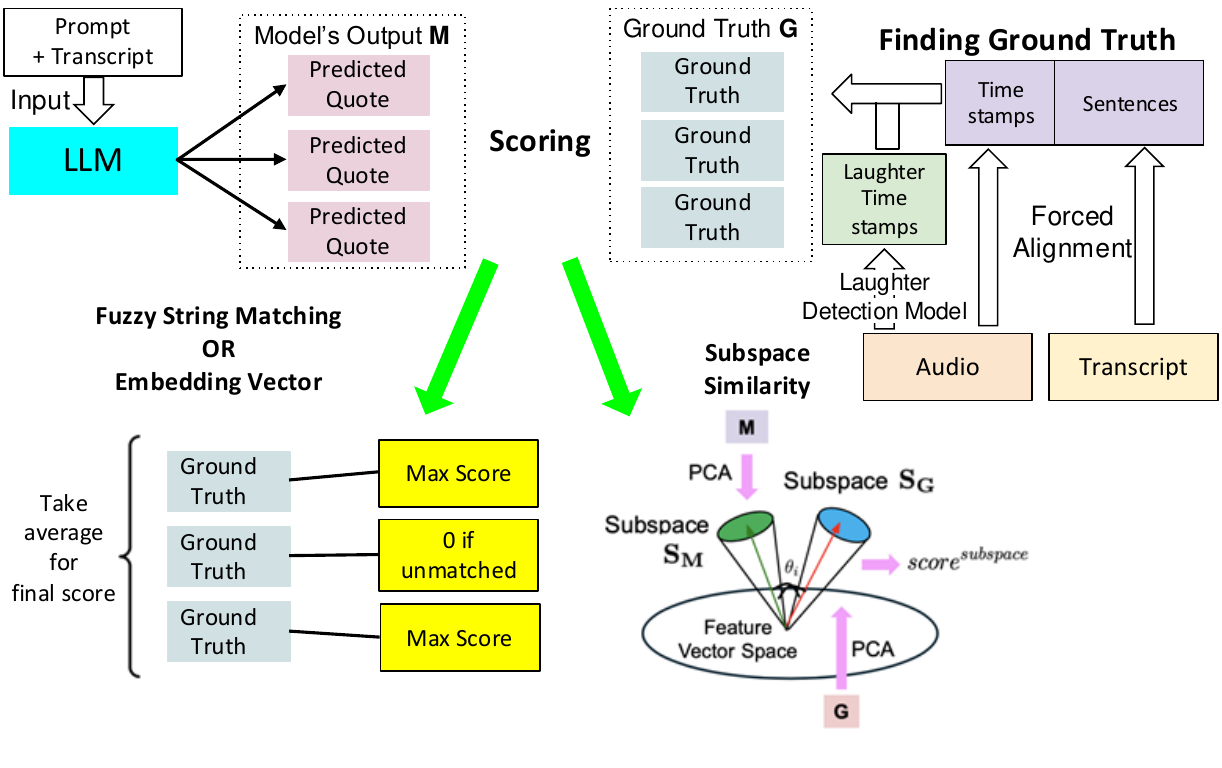}
    \caption{The humor detection metric evaluates a model's ability to identify funny quotes by comparing its outputs against the ground truth found through forced alignment and laughter detection. The metric offers three alternative scoring modules: 1) fuzzy string matching that assigns a score based on text similarity, 2) vector embeddings that compare semantic similarities, and 3) subspace similarity that analyzes the underlying patterns of a model on the task. Fuzzy string matching and the vector embedding modules operate under a similar scoring procedure, where the predicted quote is matched with ground truth quotes and assigned a similarity score, with unmatched quotes receiving a score of 0, and the average representing the final score. We integrate the metrics to assess a model's predictions with the the stand-up comedy quotes that made the audience laugh. Only one of these three modules is selected and used to generate the final metric score for evaluation.}
    \label{fig:large}
\end{figure*}

\section{Related Work}
\subsection{Computational Humor and Humor Theory} 
Humor is a widely recognized but conceptually complex phenomenon, with psychologists disagreeing on its precise definition. It encompasses three distinct constructs: sense of humor (an individual's tendency to laugh or amuse others), comedy (a stimulus that elicits laughter and amusement), and humor appreciation (the psychological response to humor). Collectively, these constructs form what we refer to as humor. Additionally, some researched describe humor as a subjective psychological reaction to comedic stimuli~\cite{warren2021what}. Through a linguistic lens, three widely recognized theories explain the phenomena of humor: the Superiority Theory, humor arises from feeling superior to others; the Relief Theory, humor releases psychological tension; and the Incongruity Theory, humor stems from the sudden violation of expectations~\cite{morreall2020philosophy}. A common task in computational humor is humor detection, identifying whether a given text or media is intended to be funny. \citet{bertero-fung-2016-deep} explore various classification algorithms to detect punchlines in the TV sitcom \textit{The Big Bang Theory} and \citet{purandare-litman-2006-humor} examine humor recognition in the TV show \textit{Friends}, employing acoustic-prosodic and linguistic features for analysis. However, both studies rely on artificial laughter rather than authentic audience reactions. \citet{Platow2005} argues that canned laughter functions as a prompt to engage viewers and bolster weaker jokes, while real audience laughter serves as a more reliable indicator of natural humor, providing an accurate reflection of comedic effectiveness. The UR-FUNNY dataset avoids artificial laughter by using TED talks in order to provide an authentic representation of humor~\cite{Hasan2019}. Stand-up comedy, with its immediate audience feedback, offers a unique advantage for humor research, as it mirrors the Incongruity Theory where comedians create an expectation through a set-up and subvert it with the punchline~\cite{amin-burghardt-2020-survey}. \citet{mittal2021so}'s Open Mic dataset of stand-up performances was used to train models to assign a "funniness" score to script segments validated by human annotators.

\subsection{LLM’s in Humor Detection} In computational humor, there is a growing interest in evaluating the humor detection capabilities of LLMs. Research in this area has explored the ability of a model to assess the funniness of jokes, with findings indicating that ChatGPT can recognize humor when prompted, though its evaluation was limited to a set of top jokes~\cite{jentzsch2023chatgpt}. Subsequent tests with a larger set of comedic content showed that zero-shot prompting resulted in ChatGPT’s humor ratings closely aligning with those of human evaluators~\cite{Goes2023}. \citet{baranov-etal-2023-told} examined humor detection across various comedic datasets using both fine-tuned models and two LLMs, ChatGPT and Flan-UL2, as zero-shot classifiers. While these models achieved high results, they did not outperform fine-tuned models. Crowd Score was introduced to classify jokes using LLMs as AI judges, by providing a personality profile with zero-shot prompting~\cite{goes2022crowd}. To the best of our knowledge, there has been no research focusing on statistical metrics for evaluating the accuracy of zero-shot settings in LLMs for detecting humor.

\subsection{Subspaces in NLP} Using word subspaces for text representation and the mutual subspace method framework for text classification extends on using word embeddings like word2vec ~\cite{shimomoto2018textclassificationbasedword}. While embeddings represent word semantics as vectors, word subspaces capture the intrinsic variability of features in a set of word vectors in order to preserve semantic relationships. Subspace representations leverage the geometric structure of embeddings to address the challenge of effective text classification with limited training data~\cite{Shimomoto2024Subspace}.  

\section{Methodology}
In this section, we will explain our proposed metric and its mathematical details. It is crucial to consider a metric that can evaluate the model's understanding of what makes a text humorous, despite the broad and subjective nature of humor.

\subsection{Humor Detection Metric} 
Our metric utilizes three alternative approaches for  scoring that capture the similarity of the model's answers to the ground truth. As shown in Figure~\ref{fig:large}, the model's score is computed in the following:
\begin{enumerate}
    \item The model is prompted to extract humorous quotes from a stand-up comedian’s transcript. These quotes are stored as a list of strings, with \( M = \{m_1, \dots, m_n\} \) being the set of quotes predicted to be funny by the model for a specific transcript. 
    \item The ground truth is determined from the transcript using a laughter detection model~\cite{gillicklaughterdetection} that extracts laughter time stamps from the accompanying audio recording~\cite{mittal2021so}. Forced alignment allows for a mapping between sentences in the transcript and laughter time frames. Thus, let \( G = \{g_1, \dots, g_k\} \) be the set of ground truths for the same transcript.
    \item We calculate how close \( M\) is to \( G\) by offering a scoring module that allows for the use of either fuzzy string matching, sentence embeddings, or subspace similarity. 
\end{enumerate}

The following contain explanations of each scoring module.

\subsection{Fuzzy String Matching Module}
Fuzzy string matching provides a straightforward approach for comparing text using Levenshtein distance~\cite{1966SPhD...10..707L}. For a given transcript, a similarity score, $\mathrm{s}^{fuzzy}$, between every model output and ground truth is stored in a similarity matrix $\mathbf{S}^{fuzzy} \in {[0, 1]}^{n \times k}$: 

\begin{equation}
    \mathbf{S}^{fuzzy}_{ij} = \mathrm{s}^{fuzzy}(m_i, g_j).
\end{equation} 

% $\mathrm{s}_{fuzzy}(m_i, g_j)$

Ideally, it is clear that the perfect score resembles an identity matrix, but in practice a ground truth can be matched with more than one prediction or to none. Therefore, the highest similarity score is selected for each ground truth to form a matrix that holds the best matches. In order to find the closest match, the maximum value is taken: 

\begin{equation}
t_j = \max_{m_i \in M} \mathbf{S}^{fuzzy}_{ij}.
\end{equation}

Notice that if a ground truth was not matched to any model output, $t_j$ is automatically assigned a score of 0. In the case of overgenerating quotes, which can be used as a tactic to exploit the metric, a penalty \(p\) is applied if the number of predictions \(n\) exceeds the number of ground truths \(k\):  
\begin{equation}
    p = \max(n - k, 0).
\end{equation}
The final score is adjusted with the penalty and a scaling factor, \( \alpha = 0.1 \), and the average score is computed for the transcript:

\begin{equation}
    score^{fuzzy} = \max\bigg(\frac{1}{k} \sum_{j=1}^{k} t_j - \alpha p, 0\bigg).
\end{equation}

\subsection{Vector Embedding Module}
In the second module, we switch to using sentence embeddings that better reflect context and meaning. In some cases, LLMs may generate non-compliant responses in which the output would be an explanation of the humor rather than a direct quote. Since fuzzy string matching purely focuses on character-level changes, like insertions or deletions, it fails to capture the semantic nuances, and therefore would heavily penalize the model’s predictions. \citet{yuan2021bartscoreevaluatinggeneratedtext} introduced BARTSCORE, a metric to evaluate the accuracy and effectiveness of generated text using BART, an encoder-decoder based model. We take a similar approach by using an embedding model from Sentence Transformers~\cite{reimers-2019-sentence-bert}, to apply a more flexible measure of similarity emphasizing the essence of a text. 

The similarity score, $\mathrm{s}^{embed}$, is now calculated using vectors of the quotes from \( M\) and \(G\):

\begin{equation}
    \mathbf{S}^{embed}_{ij} = \mathrm{s}^{embed}(\mathbf{m}_i, \mathbf{g}_j),
\end{equation}
where $\mathbf{m}_i$ and $\mathbf{g}_j$ are the vector representations of the model's predicted quote and ground truth quote that are currently being evaluated. The penalty and average are handled the same way as in the fuzzy string matching module to produce $score^{embed}$. 

\subsection{Subspace Similarity Module}
Fuzzy string matching and sentence embeddings allow us to evaluate each LLM from its output strings, but we can also conduct a deeper analysis by evaluating the LLMs feature vector space directly. With that in mind, we leverage the structural similarity between two subspaces~\cite{7053916} that can take into account the structure of the LLM feature vectors using multiple variations of instructions as input and the accompanying output for a transcript. Let \( \mathbf{M} = \begin{bmatrix}
    \mathbf{m}_1 & \mathbf{m}_2 & \dots & \mathbf{m}_n 
\end{bmatrix} \) represent the collection of model outputs and  \( \mathbf{G} = \begin{bmatrix}
    \mathbf{g}_1 & \mathbf{g}_2 & \dots & \mathbf{g}_k 
\end{bmatrix} \) represent the ground truths for each variation of instruction for a transcript. By applying PCA to the set of vectors, \( \mathbf{M}\) and \( \mathbf{G}\), respectively, 
%we obtain subspace bases $\mathbf{S}_\mathbf{M}, \mathbf{S}_\mathbf{G} \in \mathbb{R}^{d \times q}$, where $d$ is the dimension of the feature vectors and $q$ is $\min(n,k)$. 
we obtain the bases, $\mathbf{S}_\mathbf{M}$ and $\mathbf{S}_\mathbf{G} \in \mathbb{R}^{d \times q}$ of subspaces, ${\mathcal{S}}_M$ and ${\mathcal{S}}_G$, where $d$ is the dimension of the feature vectors and $q$ is the dimension of the subspaces. We calculate the SVD, ${\mathbf{S}_\mathbf{M}}^{\top}{\mathbf{S}_\mathbf{G}} = {\mathbf{U}}{\Sigma} {\mathbf{V}}^{\top}$, where $diag({\Sigma}) = (\kappa_{1},\ldots,\kappa_{q})$, $\kappa_{1} \geq \ldots \geq \kappa_{q}$, represents the set of singular values, which are the cosines of the canonical angles $\theta_i$. The similarity can then be defined
\begin{equation}
    {score}^{subspace} = \frac{1}{r}\sum_{i=1}^{r} \kappa_i^2,
    \label{eq:subspace_sim}
\end{equation}
where $r$ is the number of canonical angles used for score calculation.

By using subspaces, our metric allows us to simulate variations of the prompt while reducing penalization for minor variations, offering a comprehensive reflection of the model’s performance. 

\section{Experiments}
In this section, we evaluate several LLMs using the proposed metric, apply prompt engineering techniques to optimize model performance, and conduct a human-machine agreement task.

\subsection{Experimental Settings} We use the Open Mic dataset~\cite{mittal2021so}, which provides both audio and transcripts for several stand-up performances. To create a fair comparison, we randomly selected 51 transcripts with an average word length of 270 words and length of 106 seconds. We prompt each model with a transcript and the following instruction:

\begin{prompt}[Standard Humor Detection Prompt]
    Extract the key humorous lines and punchlines for this stand-up comedy transcript. Focus on the quotes highlighting the main comedic moments. List of quotes:
    \label{prompt:standard-prompt}
\end{prompt}

The model outputs a list of quotes that it found humorous. All experiments ran in less than a day. 

\subsection{Model Comparison} 
\begin{figure}[tb]
    \centering
\includegraphics[width=0.4\textwidth]{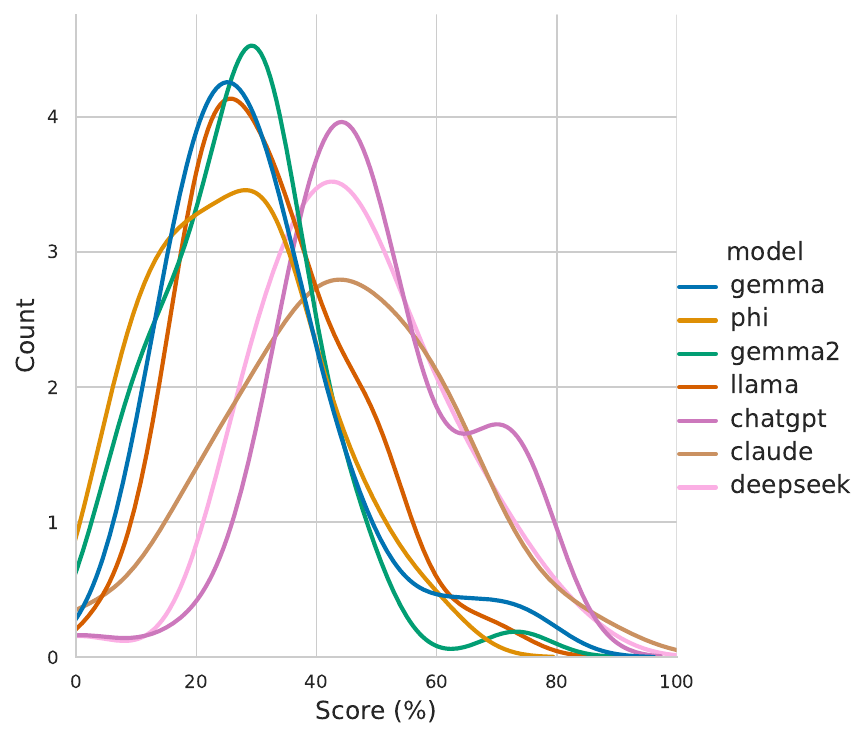}
    \caption{Distribution of scores with fuzzy string matching across several LLMs among 51 transcripts.}
    \label{fig:model-scores}
    
    \vspace{0.5cm} 
    \captionsetup{type=table} 
    \begin{tabular}{cccc}
    \hline\hline
        \textbf{Model} & \textbf{Fuzzy} & \textbf{Embed} & \textbf{Sub}\\
    \hline
        Gemma 2b-it & 30.1 & 30.0 & 55.7\\
        Gemma 2 9b-it & 35.2 & 35.9 & 35.9\\
        Phi 3-Mini 3.8b-it & 26.4 & 25.8 & 33.6\\
        Llama 3 8b-it & 31.9 & 33.8 & 38.4\\
        ChatGPT-4o & 48.9 & 25.4 & --  \\
        Claude 3.5 Sonnet & 43.4 &  46.9 & -- \\
        DeepSeek-V3 & 46 &  51.6 & -- \\
    \hline
    \end{tabular}
    \caption{Scores (\%) across models against all three metric modules using 51 transcripts.}
    \label{tab:all_scores}
\end{figure}

We evaluate various models using Prompt~\ref{prompt:standard-prompt} to gain deeper insight into our metric's assessments and explore the ability of LLMs in detecting humor. We use the instruct versions of Google’s Gemma with 2-billion parameters, Google's Gemma 2 with 9-billion parameters, Meta's Llama 3.1 with 8-billion parameters, and Microsoft's Phi 3-Mini with 3.8-billion parameters. We continue experimentation with OpenAI’s ChatGPT-4o, Anthropic’s Claude 3.5 Sonnet \footnote{Experiments were conducted in December 2024}, and DeepSeek-V3 \footnote{Experiment was conducted in January 2025} known for their advanced ability to engage in human-like interactions. These models have been employed in various studies, particularly in joke detection, generation, and evaluation using many-shot prompting~\cite{jentzsch2023chatgpt,deepseekai2024deepseekv3technicalreport}. Figure~\ref{fig:model-scores} shows the average scores for each model found with fuzzy string matching and Table~\ref{tab:all_scores} shows results with all modules. Interestingly, ChatGPT performs well using fuzzy string matching but exhibits a significant decline in performance with semantic similarity metrics. This discrepancy suggests that while ChatGPT excels in identifying quotes with high lexical similarity, it struggles to capture deeper semantic relationships.

\begin{figure}[tb]
    \centering
\includegraphics[width=0.35\textwidth]{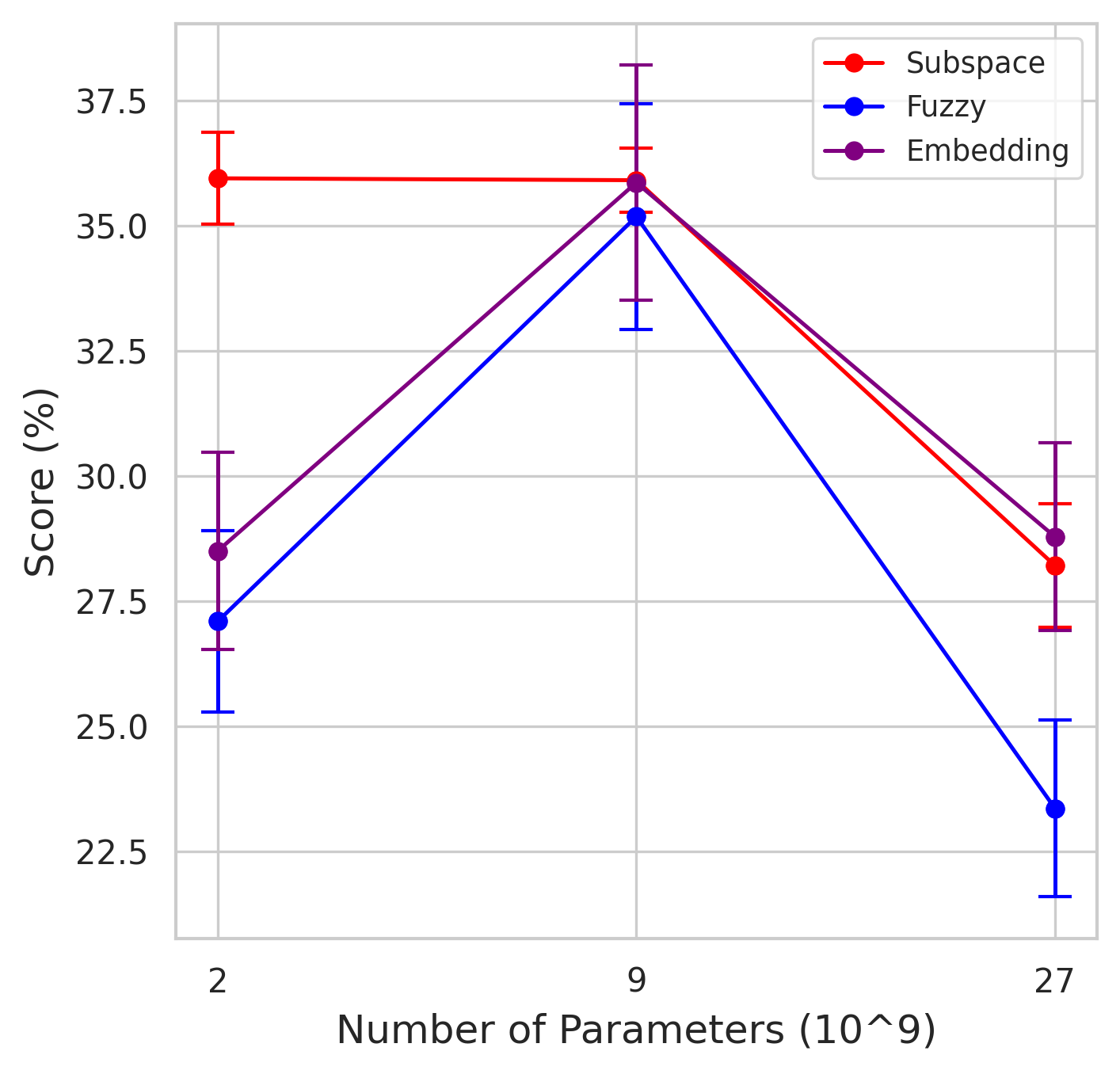}
    \caption{Evaluation of the Gemma 2-it family among model sizes using all three modules.}
    \label{fig:subspace_metric}
\end{figure}

Given Gemma 2's high performance, we further evaluate the model across varying model sizes among all scoring modules. The results in Figure~\ref{fig:subspace_metric} suggest a potential relationship between the nature of the task and the architecture of the model. In general, models with higher parameter configurations tend to succeed in logical tasks, as opposed to subjective tasks~\cite{chen2024rolesmallmodelsllm}. Additionally, the 27-b parameter model exhibited more instances of misaligned outputs to the prompt, where it not only listed a quote but provided an explanation of why the quote was funny. Thus, this difficulty of capturing humor's nuances may account for the model's low scores. 

% The Pearson correlation for the Gemma 2 9-b configuration for each module is displayed in Table \ref{tab:spear}. Fuzzy string matching and sentence embedding modules show a strong positive correlation between evaluations. In contrast, the subspace module combinations present a weak relationship. Subspaces use multiple variations of instructions rather than a single prompt. Consequently, the weaker association with the two other modules may be reflective of this methodological difference.

\subsection{Prompt Engineering}	A model’s performance on a task can be heavily dependent on the input they receive. Prompt engineering focuses on crafting inputs to elicit a desired response. For humor detection, we focus on maximizing the model's ability to retrieve humorous quotes and measure the performance throughout various prompt designs. All evaluations were done using the fuzzy string matching module.

% \begin{table}[tb]
%     \centering
%     \begin{tabular}{cc}
%     \hline\hline
%         \textbf{Modules} & \textbf{Correlation}\\
%          \hline
%        Fuzzy $\times$ Embed  & 0.91 \\
%         Fuzzy $\times$ Subspace & 0.06 \\
%         Embed $\times$ Subspace & 0.04\\
%         \hline
%     \end{tabular}
%     \caption{Pearson correlation among all transcripts between scoring modules for Gemma 2 with 9-b parameters.}
%     \label{tab:spear}
% \end{table}

In order to generate a list of prompts, we provided ChatGPT with a transcript and ground truth and asked, “\textit{If I wanted a model to extract this list of quotes from the following stand-up comedy transcript, what would the best instruction be?}”. The results are shown at Prompts~\ref{prompt:chatgpt-1}, \ref{prompt:chatgpt-2}~and~\ref{prompt:chatgpt-3}.

\begin{prompt} \label{prompt:chatgpt-1}
    When performed in front of a live audience, which jokes do you think made the audience laugh?
\end{prompt}

\begin{prompt} \label{prompt:chatgpt-2}
    What are the funniest punchlines from the transcript?
\end{prompt}

\begin{prompt} \label{prompt:chatgpt-3}
    Analyze the transcript and extract the quotes that are most likely to have made the audience laugh.
\end{prompt}

An assessment of Gemma 2b-instruct can be seen in Table~\ref{tab:PromptEngineeringScores}. Prompt~\ref{prompt:chatgpt-1} received the highest score and the remaining prompts had no positive change in performance. 

A popular technique for prompt engineering is \textit{The Persona Pattern}, where the model is given a role that guides it into focusing on specific details when generating an output \cite{white2023promptpatterncatalogenhance}. We sought to examine how the scores of Gemma 2b-instruct would be affected across personas. First, the model was assigned three distinct roles: a comedian, a comedy fan, and a comedy critic. The same instructional prompt (\ref{prompt:persona-prompt}) was employed across all roles. Table~\ref{tab:PromptEngineeringScores} indicates that personas do not make relevant changes to the scores.

\begin{prompt}[Persona Pattern Prompt]\label{prompt:persona-prompt}
    Pretend that you are a [insert role] reading the following stand-up comedy transcript.
\end{prompt}
                        
Although previous persona adoption showed no improvement, \citet{goes2022crowd}'s success in evaluating jokes with roles that specialize in categories of humor inspired a similar approach in this study. We instructed Gemma 2 9b-instruct to embody an individual who enjoys a specific type of humor following the template at Prompt~\ref{prompt:preference-prompt}. However, as seen in Table~\ref{tab:HTScores}, the prompt with no specialization received the highest score, suggesting that humor-specialized prompts do not enhance performance.

\begin{prompt}[Humor Preference Prompt]\label{prompt:preference-prompt}
    You are a person who enjoys [insert humor type] humor.
\end{prompt}

We prompt the model with the comedian whose transcript it was analyzing. This was implemented using Gemma 2b-instruct and ChatGPT-4o, which has previously showed the capability for celebrity impersonation~\cite{yokoyama-etal-2024-aggregating}. Despite earlier success in mimicking famous individuals, Table~\ref{tab:comScores} shows no improvements in humor detection, despite the comedians being quite well known.

\begin{prompt}[Audience Demographic Prompt]\label{prompt:demoe-prompt}
    Pretend you are a [insert gender/race/age].
\end{prompt}

Prompt engineering has been used to target specific audience demographics \cite{choi2024proxonaleveragingllmdrivenpersonas}. In this study, we assign Gemma 2 with varying race, ages, and gender to investigate if scores change based on demographics. We assign a race of either Caucasian/White, Black/African American, Hispanic/Latino, or Asian. We chose the age ranges of teenager (13-18 years), young adult (18-34 years), adult (35-64 years), and elderly (65+). Lastly, we use a female or male persona. In Table~\ref{tab:DemoScores}, no specific demographic yields improvement compared to the baseline, but the young adult persona resulted in the closest performance, suggesting a marginal alignment with the model's inherent capabilities.

\begin{table}[tb]
\centering
 \begin{center}
  \begin{tabular} { c | c | c | c }
  \hline\hline
    \multicolumn{4} { c }{\textbf{Prompt Engineering}} \\
    \hline\hline
    Original & Prompt 1 & Prompt 2 & Prompt 3 \\
    \hline
    30.1\% & 27.4\% & 31.2\% & 28\% \\
    \hline\hline
    \multicolumn{4} { c }{\textbf{Persona Prompts}} \\
    \hline\hline
    Original & Comedian & Fan  & Critic \\
    \hline
    30.1\% & 28.7\% & 27.9\% & 30.5\% \\
    \hline\hline
\end{tabular}
  \end{center}
\caption{Average scores found using fuzzy string module for prompt engineering for Gemma 2b-instruct.}
\label{tab:PromptEngineeringScores}
% \end{table}
    \vspace{0.5cm} 
% \begin{table}[t]
\centering
 \begin{center}
  \begin{tabular} { c | c | c | c }
  \hline\hline
    \multicolumn{4} { c }{\textbf{Humor Type Prompt}} \\
    \hline\hline
    Original & Aggressive & Dark & Deprecating \\
    \hline
    35.2\% & 32.7\% & 31.2\% & 32.0\% \\
    \hline\hline
\end{tabular}
  \end{center}
\caption{Average scores found using fuzzy string module for different humor types as personas for Gemma 2 9b-instruct.}
\label{tab:HTScores}
% \end{table}
    \vspace{0.5cm} 
% \begin{table}[t]
\centering
 \begin{center}
  \begin{tabular} { c | c | c }
  \hline\hline
    \multicolumn{3} { c }{\textbf{Stand-up Comedian Persona}} \\
    \hline\hline
     & ChatGPT-4o & Gemma 2b-instruct  \\
    \hline
    Original & 50.3\% & 27.1\% \\
    \hline
    Persona & 45.2\% & 26.0\% \\
    \hline\hline
\end{tabular}
  \end{center}
\caption{Prompt engineering average scores using fuzzy string module for ChatGPT-4o and Gemma 2b-instruct when taking the role of the comedian whose transcript it was analyzing.}
\label{tab:comScores}
\end{table}

\newcolumntype{C}{>{\centering\arraybackslash}X}
\begin{table}[tb]
\centering
\begin{tabular} { c | c | c | c | c }
  \hline\hline
    \multicolumn{5}{c}{\textbf{Race}} \\
  \hline
    None & White & Black & Hispanic & Asian \\
  \hline
    35.2\% & 31.1\% & 28.7\% & 30\% & 26.9\% \\
  \hline\hline
    \multicolumn{5}{c}{\textbf{Age}} \\
  \hline
    None & Teen & YA & Adult & Elderly \\
  \hline
    35.2\% & 32.8\% & 34.2\% & 31.7\% & 28.7\% \\
  \hline\hline
\end{tabular}
\begin{tabularx}{\linewidth}{C|C|C}
  \multicolumn{3}{c}{\textbf{Gender}} \\
  \hline
  None & Woman & Man \\
  \hline
  35.2\% & 31.9\% & 33.3\% \\
  \hline\hline
\end{tabularx}
\caption{Average scores found using fuzzy string module for audience demographic prompt for Gemma 2b-instruct.}
\label{tab:DemoScores}
\end{table}
\begin{table}[tb]
    \centering
    \begin{tabular}{cc}
        \hline    \hline
       \textbf{Model}  & \textbf{\%}\\
        \hline
        Gemma 2b-instruct & 68.8\\
        Gemma 2 9b-instruct & 68.8\\
        Llama 3 8b-instruct & 61.1\\
        Phi3-Mini 3.8b-instruct & 66.9\\
        ChatGPT-4o & 28.7\\
        Claude 3.5 Sonnet & 65.0\\
        DeepSeek-V3 & 58.9 \\
    \hline
        \textbf{Average} & 59.9\\
        \hline
    \end{tabular}
    \caption{Agreement scores between human evaluators and LLMs.}
    \label{tab:hm}
\end{table}

\subsection{Human-Machine Agreement}
Human evaluation remains one of the most valuable methods for assessing LLM performance, especially when examining a subjective output like humor. Thus, we asked 11 participants to perform the same task as the models on 6 transcripts from well-known comedians. The evaluators were naive raters across various cultural backgrounds, all within an age range of 20 to 30 years.

\begin{table}[tb]
    \centering
    \begin{tabular}{c c}
    \hline\hline
    \textbf{Transcript }& \textbf{\%} \\
    \hline
        Ali Wong & 83.7 \\
        Anthony Jeselnik & 90.1\\
        Hasan Minhaj & 85.4\\
        Jimmy Yang & 87.0\\
        Joe List & 88.5\\
        John Mulaney & 85.7\\
    \hline
        \textbf{Average} & 86.7\\
    \hline
    \end{tabular}
    \caption{Agreement scores between the human evaluators on a specific comedian's transcript.}
    \label{tab:human_an}
\end{table}

Following the approach of \citet{hada2024largelanguagemodelbasedevaluators}, we compute the agreement between evaluators using Percentage Agreement (PA). Each person received the 6 transcripts, split into sentences, and was asked to mark each as funny or not. The scores in Table~\ref{tab:human_an} indicate that humans achieved a relatively high PA across all transcripts, with an average of 86.7\%. Even though participants were generally able to identify the same quotes, the absence of a perfect consensus emphasizes the subjectivity of the task. It is important to note that the PA could be influenced by similar age ranges, leading to shared cultural references and senses of humor, potentially narrowing the diversity of interpretations.

We use the fuzzy string matching module to evaluate human answers against the ground truth. This revealed that humans receive a score of 40.7\%. Interestingly, leading models ChatGPT, Claude, and DeepSeek, when measured with the same module, outperform humans. This disparity may arise because LLMs are inherently optimized for text-based tasks, focusing on linguistic and semantic cues without needing situational context. \citet{mohamed2020incongruity} argues that humor in stand-up comedy often stems from incongruity, relying less on a performer’s stage persona and more on linguistic mechanisms. In the absence of theatrical embellishments, models excel at language-centric tasks and are particularly adept at identifying puns and wordplay. In contrast, humans often rely on elements such as delivery, tone, and audience reactions, which are absent in written transcripts, potentially limiting their ability on the task. We hypothesize that the scores for humans may differ if the evaluators were tasked with focusing on textual properties rather than general context.

The human-machine agreement rate between each model and humans was found with PA. For humans, a quote was funny if majority of raters voted on it. The scores can be found in Table \ref{tab:hm}. 

Gemma 1 and 2 have the highest agreement rates, meaning that humans and these models agreed most on the funniness of a quote. The average agreement rate reaches 59.9\%, suggesting that while there is a notable level of alignment in humor detection, pinpointing the same quotes proves to be difficult. It is interesting to note that Gemma 2 and humans received similar scores with the metric's evaluation, suggesting a high level of similarity in how the model and humans assessed humor in a text-based format. Despite receiving a high score with the metric, ChatGPT has the lowest agreement rate, demonstrating that the agreement rate and metric scores do not have to match. ChatGPT's ability surpassing humans on the task is unrelated to the agreement rate. 

\section{Conclusion}
In this work, we introduce a novel humor detection metric designed to score a model's output in relation to the ground truth of a text. The metric uses a scoring module in which the model can be evaluated using fuzzy string matching, sentence embeddings, or subspace similarity. We use a stand-up comedy dataset that offers unique narratives crafted with punchlines to maximize audience laughter. The ground truth is derived from laughter during the performance in which the entire atmosphere is conducive to comedy, emphasizing the limitations of text-based analysis. The task of identifying humor in a transcript appears to be a challenge, with even leading models, such as ChatGPT, Claude, and DeepSeek, barely receiving scores over 50\%. However, this difficulty is also evident among humans, who  only received a score of 40.7\% when assessed with the metric, revealing that leading models can outperform humans on the task. 

In the future, we aim to apply the metric to evaluate a model's predicted quotes in a format distinct from text. Stand-up comedy is heavily influenced by elements not captured in written transcripts. We hypothesize that if a model were to extract quotes from a performance with muted laughter, the nature of the output would differ substantially. Moreover, this approach raises questions about the perception of humor among humans when they view stand-up without background laughter. By exploring live comedy performances, we hope to deploy our metric for humor detection on stand-up comedy videos.

\section{Limitations}
This study presents some limitations regarding the calculation of ground truth and the nature of humor analysis. First, the ground truth is derived from audio recordings where laughter is marked using timestamps. Since we assume that the sentence preceding the laughter is the humorous one, there is a possibility that the most humorous part of the joke was not accurately captured. Although we accounted for potential delays in laughter, some reactions may have been misattributed. Second, the ground truth does not differentiate between varying magnitudes of laughter. We used a laughter detection model with a minimum laughter length of 0.2 seconds and a minimum probability threshold of 0.5 (default values)~\cite{gillicklaughterdetection}, which may have resulted in some laughter being missed. Thus, jokes that elicited subtler audience reactions might not have been accounted for. Lastly, our study relies on a text-based analysis of humor, which is a clear limitation when evaluating performances originally designed for live delivery. Future research could explore how incorporating non-textual elements—such as tone, timing, and body language—affects humor perception for both human evaluators and language models.

\section{Ethical Statement}
In this work, we use stand-up comedy audio recordings and transcripts, which may contain humor that some may find offensive or politically incorrect. The content was analyzed solely for research purposes, without endorsement of any particular viewpoint.

\section*{Acknowledgement}
This work was supported by JSPS KAKENHI Grant Number JP23K28117.

%%%%%%%%% REFERENCES.

\end{document}